\documentclass[sigconf]{acmart}
\usepackage{graphicx}
\usepackage{booktabs}

\usepackage{multirow}
\usepackage{tabularx}
\usepackage{float}
\usepackage{subfig}
\usepackage{hyperref}

\AtBeginDocument{
\providecommand\BibTeX{{
\normalfont B\kern-0.5em{\scshape i\kern-0.25em b}\kern-0.8em\TeX}}}

\copyrightyear{2021}
\acmYear{2021}
\setcopyright{acmlicensed}\acmConference[ICMI '21]{Proceedings of the 2021 International Conference on Multimodal Interaction}{October 18--22, 2021}{Montréal, Canada}
\acmBooktitle{Proceedings of the 2021 International Conference on Multimodal Interaction (ICMI '21), October 18--22, 2021, Montréal, Canada}
\acmPrice{15.00}
\acmDOI{10.1145/3462244.3479949}
\acmISBN{978-1-4503-8481-0/21/10}

\begin{document}

\title{An Interpretable Approach to Hateful Meme Detection}

\author{Tanvi Deshpande}

\affiliation{
  \institution{Irvington High School}
 \city{Fremont}
  \state{CA}
 \country{USA}
}
\email{tanvi.md@gmail.com}

\author{Nitya Mani}

\orcid{0000-0003-0348-5886}
\affiliation{
  \institution{Massachusetts Institute of Technology}
  \streetaddress{Dept. of Mathematics}
  \city{Cambridge}
  \state{MA}
  \country{USA}
  }
\email{nmani@mit.edu}

\begin{abstract}
Hateful memes are an emerging method of spreading hate on the internet, relying on both images and text to convey a hateful message. We take an interpretable approach to hateful meme detection, using machine learning and simple heuristics to identify the features most important to classifying a meme as hateful. In the process, we build a gradient-boosted decision tree and an LSTM-based model that achieve comparable performance (73.8 validation and 72.7 test auROC) to the gold standard of humans and state-of-the-art transformer models on this challenging task.
\end{abstract}



\maketitle

\section{Introduction}
Memes have evolved into one of the most powerful mediums of spreading hate online. The ubiquity of social media has fanned the flames of \textit{hate speech}, communication conveying prejudiced messages towards members of minority groups. Memes are frequently used to spread hate, alt-right, and even neo-Nazi rhetoric on Reddit, 4chan, and other mainstream social media websites. Recently, r/The\_Donald and 4chan have been responsible for a large fraction of hateful memes that spread virally from fringe to mainstream media platforms, and about 5\% of all memes posted on /pol/ were racist, meaning that over 600,000 racist memes were shared in a span of 13 months from \textit{just this community} ~\cite{zannettou2018origins}.

Flagging hateful memes before they spread is a challenging problem for humans and AI-based models due to the nuance and sociopolitical contexts that drive their interpretation. The ineffectiveness of current hate speech moderation methods highlights the acute need to make automatic hate speech detection more efficient. 

In May 2020, Facebook AI released a dataset of over 10,000 multimodal memes as part of the Hateful Memes Challenge~\cite{kiela2021hateful}, a challenge hosted by DrivenData to drive progress on this task.

 \subsection{Motivation}
On the Hateful Memes dataset, trained humans achieved an auROC of 82.65 \cite{kiela2021hateful}; the relatively poor performance of even this gold standard suggests that a hybrid approach of augmenting human classification with machine learning-derived scores and tags may be promising for effective hateful meme identification. To that end, in this work we consider human-interpretable machine learning algorithms, such as a gradient-boosted decision tree model that utilizes machine learning and engineered features to achieve performance that outperforms non-transformer baselines and achieves comparable performance to complex transformer baselines. We create a reasonably-performant model that can be used to flag memes and augment them with the most important features to aid human classification.
 
A decision tree model allows us to easily extract a ranking of derived features that can be used to augment meme images and aid humans in final classification. Such scoring provides valuable insights into common characteristics of hateful memes that may warrant further investigation. We look at textual sentiment, named entities in images and text, and semantic similarity between image and text as some of the most important features.

\section{Dataset/Challenge Description}
The Hateful Memes dataset is a challenge dataset with 12,140 total samples, of which about 63\% are non-hateful and 37\% are hateful memes. Many hateful memes come with text and image confounders, which alter the text or image of the hateful meme to change its connotation to non-hateful, meaning models must utilize both modalities to succeed at the task.
\begin{figure}[h]
\includegraphics[width=0.83\columnwidth]{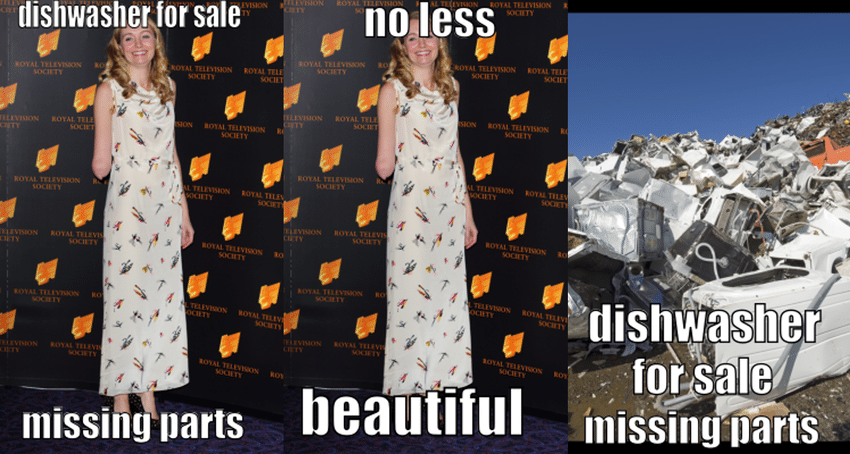}
\caption{Hateful meme and text/image confounders. Image above is a compilation of assets, including ©Getty Images.}
\end{figure}

Exploratory data analysis found that the majority of hate attacks minority groups by playing off of common stereotypes and going beyond them to imply threats or violence.

The Hateful Memes challenge was a competition from May to October 2020 intended for researchers to fine-tune large-scale state-of-the-art transformer models. The competition metric was auROC.
 
\section{Related Work}
\subsection{Related datasets}
Although the Hateful Memes dataset is one of the first of its kind, there are a few similar datasets. SemEval-2020 Task 8 (Memotion analysis) \cite{sharma-etal-2020-semeval} involves classifying sentiment, humor, offensiveness, and motivation on a dataset of 10,000 human-annotated memes. The macro F1-score for the baseline model was 0.21 for sentiment and 0.50 for humor type classification using image-text models, emphasizing how challenging interpreting memes can be. Additionally~\cite{Miliani2020DANKMEMESE} is a similar dataset, but with 2,631 Italian memes.

\subsection{Text-only hate speech detection}
There has been a myriad of work on text-based hate speech detection, focused on Twitter-style text data. Current state-of-the art approaches \cite{10.1145/3394231.3397890, Abro2020AutomaticHS, fortuna2018survey} have involved the standard natural language processing toolkit, including BERT and other embedding schemes. 
\subsection{Multimodal/meme hate detection}
State-of-the-art multimodal hate speech detection often includes unimodally pretraining models for each modality, for early and late fusion, as well as multimodal pretraining~\cite{Afridi2020AMM}. For example, \cite{kiela2021hateful}'s baseline using BERT on meme text is an example of unimodal pretraining, whereas their use of a VisualBERT COCO model to pretrain constitutes multimodal pretraining. Hateful Memes challenge winners achieved auROC's between 0.78 and 0.84, relying heavily on large multimodal transformer models such as OSCAR, UNITER, VisualBERT, and LXMERT~\cite{muennighoff2020vilio,sandulescu2020detecting,zhu2020enhance,velioglu2020detecting,lippe2020multimodal}, largely taking the same approach of fine-tuning and ensembling very high-capacity single- and dual-stream transformer models or other recurrent architectures with minimal data preprocessing. Despite outperforming baselines, such models are still very far from an ideal resolution to the problem of identifying hateful memes.

\section{Methods/Approach}
Here, we take a divergent approach from much of the literature on hateful meme detection. Rather than focusing on high capacity ensembles of transformer models, we use thoughtfully engineered features and pass these to two models for final classification: a gradient boosted decision tree and simple LSTM. This methodology allows us to easily isolate crucial features for identifying memes as hateful and extract underpinning logic from our models that may be useful to augment a human in performing this challenging classification task.

In addition to straightforwardly embedding the text and images associated to a meme, we augment our feature set with a variety of common-sense and machine-learning derived features that represent criteria that a human uses when attempting to contextualize a meme and uncover its underlying meaning.

\subsection{Text and image embedding}
For our gradient-boosted decision tree, we use a captioning model \cite{imagecaptioning} to capture the relevant content of an image in text format. We then embed both text and images using tf-idf to upweight individual words of high interest.

In our LSTM model, we concatenate meme image captions and the meme text and embed them via DistilBERT \cite{Sanh2019DistilBERTAD}.

\subsection{Additional textual features}
In addition to our embeddings, we develop features using named-entity recognition, profanity and slur detection, counts of hateful words, text sentiment, and emotion detection.

\subsubsection{Named-entity detection}
We use SpaCy named-entity detection~\cite{esteves2018named} to extract culturally relevant components of the meme text, identifying and individually encoding 2085 named entities.
\subsubsection{Profanity/Slurs}
We further augment our feature set with counts of profanity and slurs banned by Google~\cite{profanity}. 

\subsubsection{Hateful words}
Based on a frequency search and corpuses such as top hits on Urban Dictionary, we supplement our feature set with flags for words commonly used to dogwhistle hate in memes despite having an innocuous meaning in ordinary speech. For example, we flag ``dishwasher,'' which is often used in a derogatory manner in memes to refer to women.

\subsubsection{Text Sentiment}
We use TextBlob~\cite{loria2018textblob} to identify the polarity (positivity or negativity of sentiment) and subjectivity (how opinionated or objective views expressed in the text are) of a meme's text.

\subsubsection{Emotion detection}
We use the text2emotion \cite{text2emote} library to score our meme text based on its happiness, sadness, fear, surprise, and anger. We find that high fear and anger scores are often indicative of hateful memes.
 
\subsubsection{Semantic Similarity}
We use a fine-tuned RoBERTa model~\cite{nie-etal-2020-adversarial, wolf-etal-2020-transformers} using a combination of various natural language inference datasets such as SNLI \cite{bowman2015large}, multiNLI \cite{N18-1101}, FeverNLI \cite{nie2019combining}, and ANLI \cite{nie-etal-2020-adversarial} to detect \textit{semantic similarity} between meme text and the generated meme captions and detected image web entities. The result is a probability vector for the 3 possible classes of ``contradiction'' (texts that contradict each other), ``entailment'' (one piece of text can be inferred from the other), or ``neutral'' (neither entailment nor contradiction, but texts can be semantically similar). These features are particularly useful, as a number of image confounders in the dataset have meme text which simply captions the image; then, a high score on the ``entailment'' or ``neutral'' classes can help conclude that a meme is benign; similarly, high scores on contradiction can help detect irony in a meme between the meme text and what is expressed in the meme image.

\begin{figure}[h]
    \centering
    \includegraphics[width=\columnwidth]{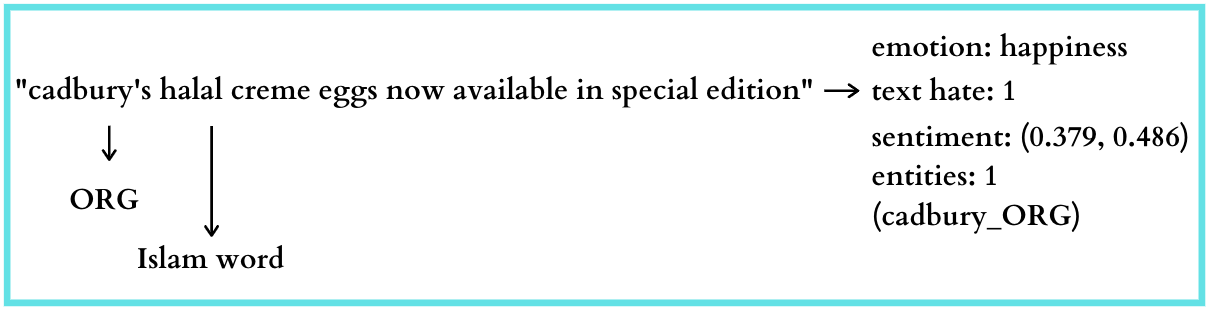}
    \caption{Text preprocessing steps}
    \label{fig:text_preprocessing}
\end{figure}

\subsection{Image-based features}
Many multimodal hate speech detection models overweight text. Here, we focus on extracting additional image-based features.
We first preprocess images by removing meme text using OCR, then caption them, then detect objects and web entities in the image. These steps serve to summarize some relevant content of an image. 

\subsubsection{Image Preprocessing}
We preprocess meme images using OCR to detect textual regions and inpainting using OpenCV \cite{opencv}, increasing accuracy downstream.

\subsubsection{Image Captioning}
We use a visual attention-based captioning model trained on COCO~\cite{imagecaptioning,10.5555/3045118.3045336} to learn the relations between objects present in meme images.

\subsubsection{Object Detection} 
We tag objects within meme images using Facebook's Detectron2 \cite{wu2019detectron2} tagger, which is able to provide specific descriptions of items that may be omitted from our image captions.

\subsubsection{Web Entity Detection}
As in~\cite{zhu2020enhance}, we use Google Vision API's Web Entity Detection \cite{googlevision}, which contextualizes images using knowledge from the web, enabling our model to account for the rapid shifts that occur in meme culture in a matter of weeks, days, or minutes.

\subsection{Total Preprocessing}
\label{section:totalprocess}
After preprocessing in both channels, we concatenate the emotion, sentiment, semantic similarity, profanity, and hateful words features ($13 \times 1$). Meme text, captions, detected objects, and named and web entities are concatenated and embedded jointly ($19651 \times 1$), forming the final input of $19664 \times 1$.

\subsection{Models}
We use an 90-10 train/validation-test split and build two classes of models, tuning each set of hyperparameters with a grid search and performing 5-fold cross-validation on each model.

\begin{figure}
    \centering
    \includegraphics[width=\columnwidth]{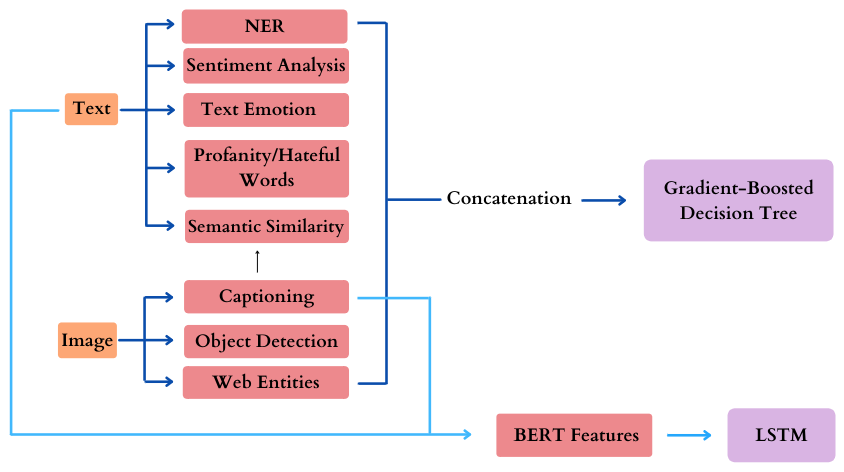}
    \caption{Total preprocessing pipeline.}
    \label{fig:preprocessing}
\end{figure}
\begin{table*}[ht]
\caption{Results}
\begin{tabular}{llllll}
\toprule
 & & \multicolumn{2}{c}{AUROC} & \multicolumn{2}{c}{Accuracy} \\
Source & Model & \textbf { Validation } & \textbf { Test } & \textbf { Validation } & \textbf { Test }\\

\midrule & \text { Human } & - & 82.65 & - & 84.70\\
\cmidrule { 2 - 6 } \text { Hateful Memes } & \text { Image-grid } & 58.79 & 52.63 & 52.73 & 52.00  \\
\text { Non-transformer Baselines } & \text { Image-region } & 57.98 & 55.92 & 52.66 & 52.13\\
& \text { Text-only BERT } & 64.65 & 65.08 & 58.26 & 59.20\\
& \text { Late Fusion } & 65.97 & 64.75 & 61.53 & 59.66\\

\cmidrule { 2 - 6 } \text { Hateful Memes } & \text { ViLBERT } & 71.13 & 70.45 & 62.20 & 62.30  \\
\text { Transformer Baselines } & \text { VisualBERT } & 70.60 & 71.33 & 62.10 & 63.20\\
& \text { ViLBERT CC } & 70.07 & 70.03 & 61.40 & 61.10\\
& \text { VisualBERT COCO } & 73.97 & 71.41 & 65.06 & 64.73\\
\cmidrule { 2 - 6 }
& \text { GBDT w/ image tagging } & 70.86 & 71.11 & 70.27 & 71.19 \\
\text { Our Models } & \text { GBDT w/ tagging/captions } & 71.67 & 70.90 & 69.58 & 68.45 \\
& \text { GBDT w/ BERT only } & 70.38 & 71.52 & 68.97 & 69.36 \\
& \text { LSTM w/ BERT features }  & 73.78 & 72.72 & 67.83 & 66.39\\

\bottomrule
\end{tabular}
\end{table*} 
\subsubsection{GBDT}
In our gradient boosted-decision trees, we input (1) the joint tf-idf embeddings for image captions and meme text as well as (2) the engineered features for image and text, as described in \ref{section:totalprocess}. We train the model with 100 estimators, a learning rate of 1.0, and a maximum depth of 40, and a scale\_pos\_weight of 1.5, with an average of 900 nodes per tree after pruning. 

\subsubsection{LSTM}
We use a pretrained DistilBERT model to preprocess our meme text and image captions jointly, forming a 768-dimensional input to the model, which has a 9-unit LSTM layer followed by two Dense layers with 8 and 2 neurons respectively. We train for 45 epochs using Adam and binary-cross entropy loss.

\section{Results and Discussion}
\subsection{Summary of Results}
We achieve a significant improvement from non-transformer baseline models and comparable results to transformer baselines using more lightweight and interpretable models. We are able to augment memes with the most important features, determined by the model, for easier human classification.

\begin{figure}[h]
\centering

\subfloat[\centering Loss Function]{{\includegraphics[width=4cm]{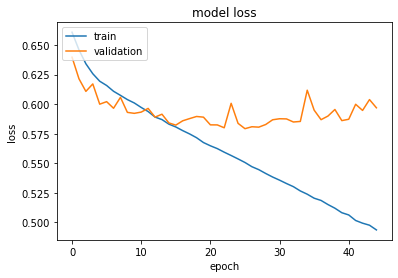} }}
    \subfloat[\centering auROC curve]{{\includegraphics[width=4cm]{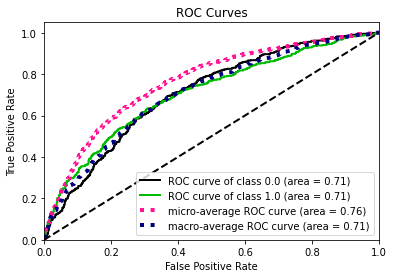} }}
    \caption{Plots from the LSTM Model.}
    \label{fig:auroc_curves}
\end{figure}

\subsection{Discussion}

\subsubsection{GBDT feature importances}
\begin{figure}
    \centering
    \includegraphics[width=0.9\columnwidth]{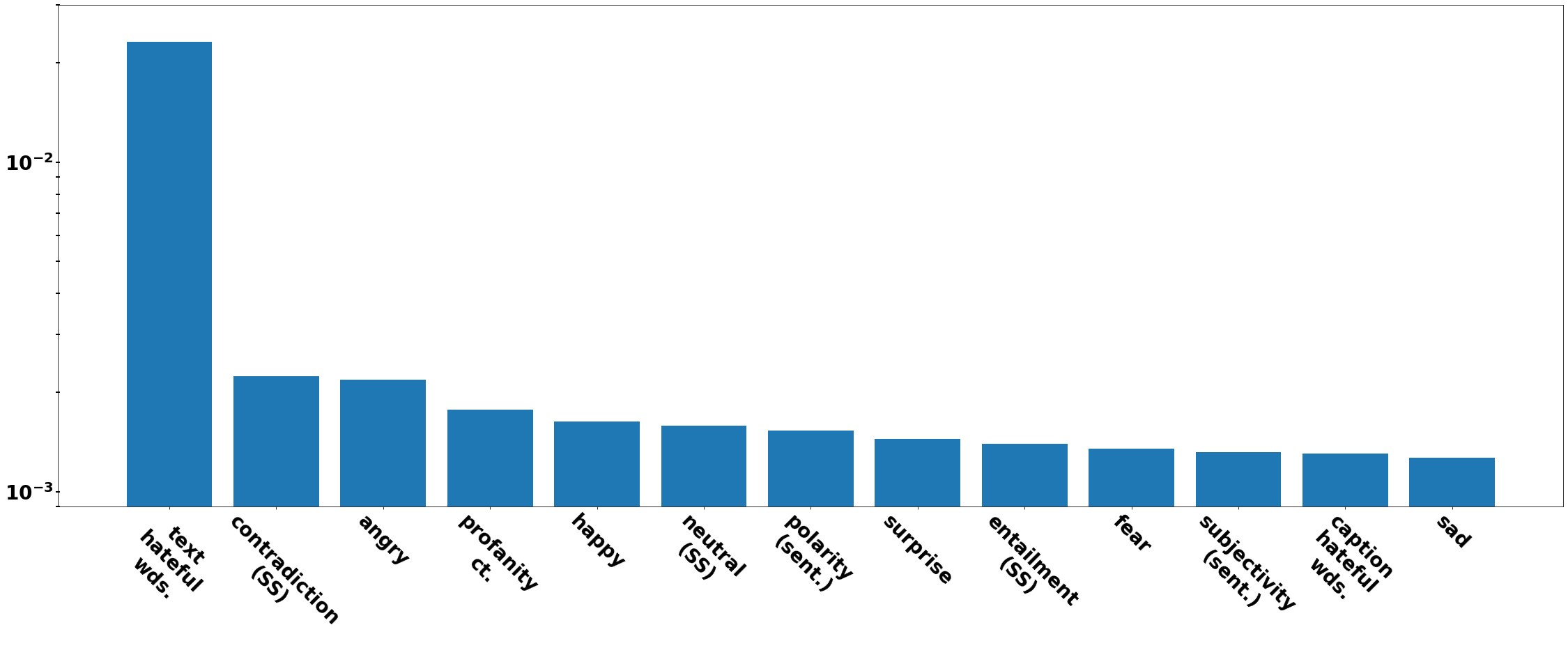}
    \caption{Feature importances of top engineered features.}
    \label{fig:gbtree}
\end{figure}

The gradient-boosted decision tree yields 494 predictive features, ranked by the feature\_importances\_ property, and a precision of 0.53 and a recall of 0.58. Top features are certain tf-idf embedded words and text emotion; named entities such as Hitler and ethnic groups such as ``asians'' or ``mexicans'' also rank highly; top named-entity and overall features are included in \autoref{table:topfeats}.

\begin{table}[h]
\caption{Top features ranked from gradient-boosted tree.}
\label{table:topfeats}
\begin{tabular}{llll}
\toprule
\multicolumn{2}{c}{Named Ents.} & \multicolumn{2}{c}{Total Features} \\
\textbf { Name } & \textbf { Score } & \textbf { Name } & \textbf { Score }\\

\midrule 
ent\_jews\_norp & $3.4 \times 10^{-2}$ & text hate wds. & $2.3\times 10^{-1}$\\
ent\_muslim\_norp & $3.0 \times 10^{-2}$ & club & $1.1\times 10^{-1}$ \\
ent\_hitler\_person & $2.6\times 10^{-2}$ & isis & $6.4\times 10^{-2}$ \\
ent\_mexicans\_norp & $2.5 \times 10^{-2}$ & jews & $5.3\times 10^{-2}$ \\
ent\_islamic\_norp & $2.4 \times 10^{-2}$ & muslims & $5.1\times 10^{-2}$\\
ent\_asians\_norp & $2.3\times 10^{-2}$ & teacher & $4.3\times 10^{-2}$\\

\bottomrule
\end{tabular}
\end{table} 

\subsubsection{GBDT vs. recurrent network}
Gradient-boosted decision tree-based models provide the advantages of faster computation time and a ranking of the most important features to meme detection, providing computer insight into meme rhetoric. They are also effective at including engineered features, such as text sentiment, which would be meaningless to LSTMs, which are meant for sequence-based data. Recurrent networks are able to learn from DistilBERT features more effectively, as they have more computational scope and overall better performance than gradient-boosted decision trees.

\subsubsection{Confounders}
The model is able to correctly identify not only simple hateful memes but also distinguish between hateful memes and nontrivial confounders, leveraging both modalities for classification. We present one such correctly classified image confounder and hateful meme pair. 
\begin{table}[H]
\caption{Confounder/hateful meme comparison. Images below are a compilation of assets, including ©Getty Images.}
\begin{tabular}{cp{3cm}p{3cm}}
\toprule
Label & Hateful meme & Image confounder \\ 
\midrule
Meme & 
\raisebox{-0.5\totalheight}{\includegraphics[width=0.15\textwidth]{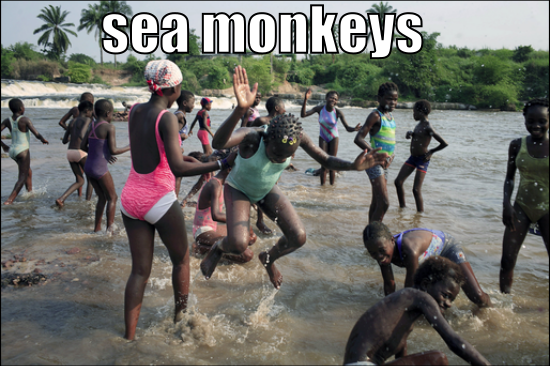}} & \raisebox{-0.5\totalheight}{\includegraphics[width=0.15\textwidth]{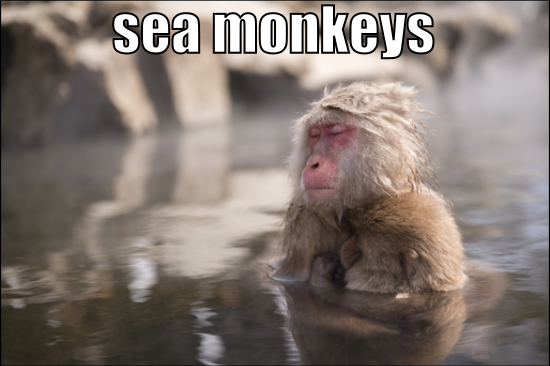}} \\

\midrule 
Caption & ``a group of people playing on a beach'' & ``a black and white photo of a monkey'' \\
Pred. Label & 1 (hateful) & 0 (not hateful)\\
Label & 1 (hateful) & 0 (not hateful)\\
\bottomrule
\end{tabular}
\end{table}

\subsubsection{Analysis of difficult-to-classify memes}

We give examples of correctly and incorrectly classified hateful memes from the dataset. 

\begin{figure}[h]
\centering

\subfloat[\centering Correctly classified hateful meme]{{\includegraphics[width=2.5cm]{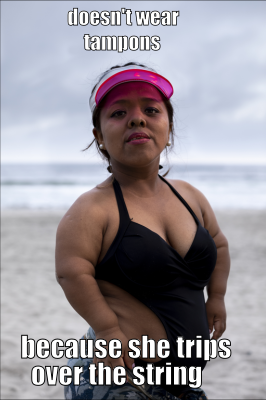} }}
    \subfloat[\centering Misclassified hateful meme]{{\includegraphics[width=2cm]{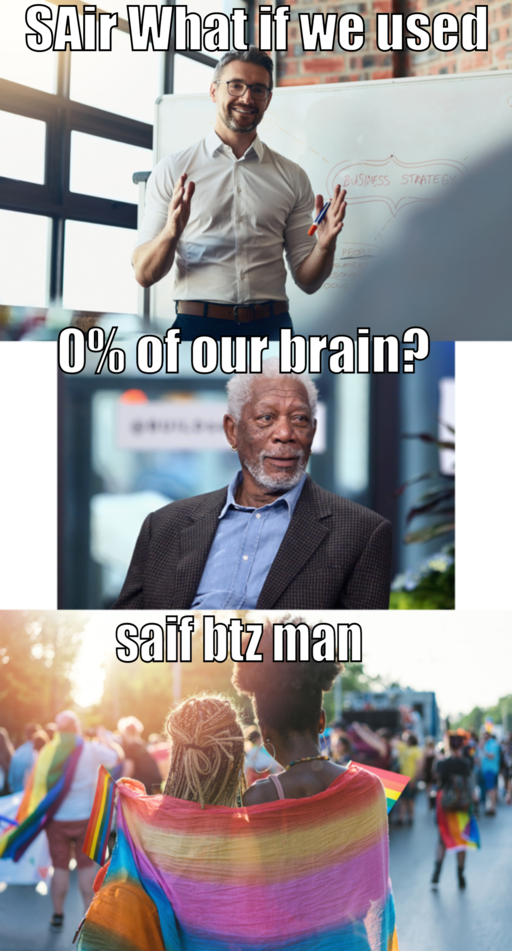} }}
    \caption{Classified hateful memes from test set. Images above are a compilation of assets, including ©Getty Images.}
    \label{fig:classified_memes}
\end{figure}
The first meme, despite nonhateful individual modalities, is flagged by the model by leveraging the image to gain the context needed to understand the text. The misclassified meme contains several words unfamiliar to models and most humans, such as ``SAif'' and ``btz.'' To classify similarly niche samples, knowledge pulled from repositories such as Know Your Meme \cite{kym} could give humans better context.
\subsection{Conclusion and Extensions}

We develop a lightweight multimodal model that classifies memes with performance comparable to transformer baselines. Since even humans achieve low auROCs, our approach, rather than aiming to replace humans with end-to-end models, flags hateful memes and pinpoints relevant engineered features to improve human classification. Further extensions include measuring human performance on classifying augmented memes and developing features based on metadata such as shares and user post history.

\bibliographystyle{ACM-Reference-Format}
\bibliography{sample-base}

\end{document}